\documentclass[conference]{IEEEtran}
\IEEEoverridecommandlockouts
\usepackage{cite}
\usepackage{amsmath,amssymb,amsfonts}
\usepackage{graphicx}
\usepackage{textcomp}
\usepackage{xcolor}
\usepackage{booktabs}
\usepackage{multirow}
\usepackage{url}

\begin{document}

\title{Degradation-Consistent Paired Training for Robust AI-Generated Image Detection
\thanks{\textsuperscript{*}Equal contribution. \textsuperscript{\dag}Corresponding author.}
\thanks{This work is supported by the Open Research Project of the State Key Laboratory of Industrial Control Technology, China (Grant No. ICT2025B70). Supported by Jiangxi Provincial Natural Science Foundation (Grant No. 20242BAB20041, Grant No. 20232BAB212006) and Hubei Provincial Natural Science Foundation of China (Grant No. 2023AFB474, Grant No. 2024AFB881). Supported by Anhui Provincial Special Project for Special Needs in Humanities and Social Sciences (Grant No. 2025AHGXSK50067) and Postgraduate Quality Engineering Project of Anhui Province (Grant No. 2024jyjxggyjY232).}
}

\author{
\IEEEauthorblockN{Zongyou Yang\textsuperscript{*}}
\IEEEauthorblockA{\textit{Department of Computer Science} \\
\textit{University College London}\\
London, United Kingdom \\
dryang0624@gmail.com}
\and
\IEEEauthorblockN{Yinghan Hou\textsuperscript{*}}
\IEEEauthorblockA{\textit{Department of Earth Science and Engineering} \\
\textit{Imperial College London}\\
London, United Kingdom \\
ghwzhyinghan@gmail.com}
\and
\IEEEauthorblockN{Xiaokun Yang\textsuperscript{\dag}}
\IEEEauthorblockA{\textit{School of Electronic Information} \\
\textit{Nanchang Institute of Technology}\\
Nanchang, China \\
yangxk@bupt.cn}
}

\maketitle

% ══════════════════════════════════════════════════════════════
\begin{abstract}
AI-generated image detectors suffer significant performance degradation under real-world image corruptions such as JPEG compression, Gaussian blur, and resolution downsampling. We observe that state-of-the-art methods, including B-Free~\cite{guillaro2025bfree}, treat degradation robustness as a byproduct of data augmentation rather than an explicit training objective. In this work, we propose Degradation-Consistent Paired Training (DCPT), a simple yet effective training strategy that explicitly enforces robustness through paired consistency constraints. For each training image, we construct a clean view and a degraded view, then impose two constraints: a feature consistency loss that minimizes the cosine distance between clean and degraded representations, and a prediction consistency loss based on symmetric KL divergence that aligns output distributions across views. DCPT adds zero additional parameters and zero inference overhead. Experiments on the Synthbuster benchmark (9 generators, 8 degradation conditions) demonstrate that DCPT improves the degraded-condition average accuracy by 9.1 percentage points compared to an identical baseline without paired training, while sacrificing only 0.9\% clean accuracy. The improvement is most pronounced under JPEG compression (+15.7\% to +17.9\%). Ablation further reveals that adding architectural components leads to overfitting on limited training data, confirming that training objective improvement is more effective than architectural augmentation for degradation robustness.
\end{abstract}


% ══════════════════════════════════════════════════════════════
\section{Introduction}

The rapid advancement of generative models---including diffusion models~\cite{rombach2022ldm,podell2023sdxl}, GANs~\cite{goodfellow2014gan}, and autoregressive generators---has made the detection of AI-generated images an urgent forensic challenge. Beyond unconditional synthesis, generative techniques have rapidly expanded into a wide spectrum of editing and content-creation tasks, such as gradient-guided image inpainting with explicit modeling of context validity~\cite{shi2026validity}, texture generation with adaptive fusion for image inpainting~\cite{shi2026texture}, and talking head video generation jointly driven by keypoints and action units for video conferencing~\cite{shi2025keypoints}. The proliferation of such generative pipelines means that AI-manipulated content reaching downstream platforms is increasingly diverse in modality, semantics, and post-processing, which in turn raises the bar for detectors to remain reliable under varied real-world degradation and compression conditions. While recent detectors achieve strong performance in controlled settings, their robustness to common image corruptions encountered in real-world deployment remains limited~\cite{gragnaniello2021gan}.

Images shared on social media platforms routinely undergo JPEG compression, resolution downscaling, and blur. These degradations can cause detector accuracy to drop by over 20 percentage points---for example, our experiments show that a DINOv2-based~\cite{oquab2024dinov2} classifier's accuracy drops from 86.2\% to 52.3\% under JPEG compression at quality factor 30. This vulnerability undermines the practical utility of otherwise strong detectors.

B-Free~\cite{guillaro2025bfree} addresses data bias through semantically aligned real-fake pair training with a DINOv2 ViT-B/14 backbone~\cite{oquab2024dinov2} with register tokens~\cite{darcet2024registers}, fine-tuned end-to-end, achieving state-of-the-art cross-generator generalization. Its \textit{inpainted++} augmentation strategy includes blur, JPEG, and scaling corruptions as data augmentation. However, the training objective remains a single cross-entropy loss---the model is never \textit{explicitly required} to maintain consistent predictions between clean and degraded versions of the same image.

We propose \textbf{Degradation-Consistent Paired Training (DCPT)}, a simple yet effective training strategy that makes degradation robustness an explicit, first-class training objective. Our key contributions are:

\begin{enumerate}
\item A \textbf{paired consistency training objective} that constructs (clean, degraded) view pairs for each training image and enforces two explicit constraints: feature-level cosine consistency and prediction-level symmetric KL divergence consistency. This adds \textbf{zero parameters} and \textbf{zero inference cost}.

\item \textbf{Empirical validation} on Synthbuster (9 generators $\times$ 8 degradation conditions) showing +9.1\% degraded average accuracy improvement, with the most dramatic gains under JPEG compression (+15.7\% to +17.9\%) at a cost of only 0.9\% clean accuracy.

\item A \textbf{revealing ablation} demonstrating that adding architectural components (frequency residual branch, +1.4M parameters) causes severe overfitting, confirming that \textbf{training objective improvement is more effective than architectural augmentation} in the low-data regime.
\end{enumerate}

% ══════════════════════════════════════════════════════════════
\section{Related Work}

\subsection{AI-Generated Image Detection}

Existing detection approaches divide into artifact-driven and representation-driven paradigms. \textbf{Artifact-driven methods} exploit generator-specific traces: Li et al.~\cite{li2020facexray} detect forgeries via blending boundaries; Corvi et al.~\cite{corvi2023diffusion} analyze spectral fingerprints unique to diffusion models; Tan et al.~\cite{tan2024npr} exploit pixel-level up-sampling artifacts; Mandelli et al.~\cite{mandelli2022orthogonal} use orthogonal training for GAN detection.

\textbf{Representation-driven methods} leverage pretrained features. Wang et al.~\cite{wang2020cnnspot} demonstrated cross-generator transfer using a ResNet-50~\cite{he2016resnet} classifier. Ojha et al.~\cite{ojha2023unifd} freeze a CLIP~\cite{radford2021clip} ViT~\cite{dosovitskiy2021vit} and train a linear probe. Tan et al.~\cite{tan2025c2pclip} inject category-common prompts into CLIP for more robust deepfake detection.

\textbf{Bias-free training.} Guillaro et al.~\cite{guillaro2025bfree} identified that detectors often learn dataset biases rather than generation artifacts. Their B-Free method generates fakes via self-conditioned inpainting with Stable Diffusion 2.1~\cite{rombach2022ldm}, ensuring semantic alignment. Using a DINOv2 ViT-B/14~\cite{oquab2024dinov2} with register tokens~\cite{darcet2024registers} fine-tuned end-to-end, B-Free achieves state-of-the-art generalization.

\subsection{Robustness to Image Degradation}

Gragnaniello et al.~\cite{gragnaniello2021gan} provided the first systematic evaluation of detectors under JPEG and blur, revealing accuracy drops exceeding 20 percentage points. The GenImage benchmark~\cite{zhu2023genimage} formalized degradation evaluation with standardized splits. Cozzolino et al.~\cite{cozzolino2024raising} analyzed the challenge of robustness to post-processing as a key open problem. B-Free's \textit{inpainted++} augmentation includes degradations but only as data diversity---not as an explicit training constraint. Our DCPT fills this fundamental gap.

\subsection{Consistency Learning and Adapters}

Our paired consistency training draws on two lines of research. \textbf{Contrastive learning} methods like SimCLR~\cite{chen2020simclr} and MoCo~\cite{he2020moco} enforce view-consistency across augmentations for self-supervised pretraining. Our approach shares this intuition but constructs clean-degraded pairs (not random augmentation pairs) and operates on supervised classification with both feature-level and prediction-level constraints. \textbf{Knowledge distillation}~\cite{hinton2015distilling} transfers knowledge via soft output matching. Our prediction consistency loss is structurally similar: the clean view acts as an implicit teacher (via stop-gradient) guiding the degraded view, but both share the same network.

For parameter-efficient adaptation, FiLM~\cite{perez2018film}, adapters~\cite{houlsby2019adapter}, and LoRA~\cite{hu2022lora} enable efficient fine-tuning. We explored a FiLM-based degradation adapter but found that architectural additions cause overfitting on limited data (Section~\ref{sec:ablation}), leading us to focus on the training objective alone.

% ══════════════════════════════════════════════════════════════
\section{Method}

\subsection{Preliminaries: B-Free}

B-Free~\cite{guillaro2025bfree} trains a binary classifier on semantically aligned (real, fake) pairs generated via self-conditioned inpainting with Stable Diffusion 2.1~\cite{rombach2022ldm}. In the original B-Free, a DINOv2 ViT-B/14~\cite{oquab2024dinov2} with register tokens~\cite{darcet2024registers} is fine-tuned end-to-end (86.5M parameters), achieving strong performance. The \textit{inpainted++} augmentation applies blur, JPEG, and scaling to increase data diversity. However, no explicit training objective enforces that the model's features or predictions should remain invariant under these corruptions.

\textbf{Our setting.} To isolate the contribution of the training objective from the backbone capacity, we adopt a \textit{frozen-backbone} setting: the DINOv2 ViT-B/14 backbone is kept frozen, and only a lightweight classifier head $g$ (two-layer MLP: $768 \rightarrow 256 \rightarrow 2$, 197K parameters) is trained. This deliberate choice ensures that any observed improvement from DCPT is attributable solely to the consistency loss, not to differences in backbone training. We extract features $\mathbf{f} = \phi(\mathbf{x}) \in \mathbb{R}^{768}$ and produce logits $\hat{y} = g(\mathbf{f})$.

\subsection{Degradation-Consistent Paired Training}
\label{sec:dcpt}

\textbf{Motivation.} Data augmentation and training objectives serve fundamentally different roles. Augmentation increases input diversity, but the model is free to ignore the degradation if it can still minimize cross-entropy using shortcuts. An explicit consistency constraint directly penalizes representation changes under degradation, regardless of whether the classification loss is satisfied.

\textbf{View construction.} For each training image $\mathbf{x}$ with label $y \in \{0, 1\}$, we construct two views:
\begin{itemize}
\item \textbf{Clean view} $\mathbf{x}_c$: original image with standard augmentation (random crop to $504 \times 504$, random horizontal flip).
\item \textbf{Degraded view} $\mathbf{x}_d$: with probability $p_{\text{deg}} = 0.5$, one of three degradation types is uniformly sampled and applied to $\mathbf{x}_c$.
\end{itemize}

The degradation types and parameter ranges are designed to match common real-world corruptions and standard evaluation protocols:
\begin{itemize}
\item JPEG compression: quality factor $q \in \{30, 50, 70, 90\}$
\item Gaussian blur: $\sigma \in \{1.0, 2.0, 3.0\}$
\item Downsampling: scale $s \in \{0.25, 0.5, 0.75\}$ (bicubic down, bilinear up to original size)
\end{itemize}

\textbf{Dual-path forward.} Both views are independently processed by the \textit{same} frozen backbone $\phi$ and classifier $g$:
\begin{equation}
\mathbf{f}_c = \phi(\mathbf{x}_c), \quad \mathbf{f}_d = \phi(\mathbf{x}_d) \quad \in \mathbb{R}^{768}
\end{equation}
\begin{equation}
\hat{y}_c = g(\mathbf{f}_c), \quad \hat{y}_d = g(\mathbf{f}_d) \quad \in \mathbb{R}^{2}
\end{equation}

The backbone and classifier are shared (not separate networks), so the consistency constraint acts directly on the model's own representations.

\textbf{Total loss.} The training objective combines four terms:
\begin{equation}
\mathcal{L} = \underbrace{\text{CE}(\hat{y}_c, y) + \text{CE}(\hat{y}_d, y)}_{\mathcal{L}_{\text{cls}}} + \lambda_f \cdot \mathcal{L}_{\text{feat}} + \lambda_p \cdot \mathcal{L}_{\text{pred}}
\label{eq:total_loss}
\end{equation}

\textbf{Feature consistency loss} $\mathcal{L}_{\text{feat}}$ enforces that backbone representations remain aligned under degradation:
\begin{equation}
\mathcal{L}_{\text{feat}} = 1 - \frac{\mathbf{f}_c \cdot \mathbf{f}_d}{\|\mathbf{f}_c\| \cdot \|\mathbf{f}_d\|}
\label{eq:feat_loss}
\end{equation}

This is the mean cosine distance, bounded in $[0, 2]$. When $\mathcal{L}_{\text{feat}} = 0$, the features are perfectly aligned regardless of degradation.

\textbf{Prediction consistency loss} $\mathcal{L}_{\text{pred}}$ aligns output probability distributions using symmetric KL divergence with stop-gradient on the clean view:
\begin{equation}
\mathcal{L}_{\text{pred}} = \text{KL}(\bar{p}_c \| p_d) + \text{KL}(p_d \| \bar{p}_c)
\label{eq:pred_loss}
\end{equation}
where $p_c = \text{softmax}(\hat{y}_c)$, $p_d = \text{softmax}(\hat{y}_d)$, and $\bar{p}_c = \text{sg}[p_c]$ denotes stop-gradient. The stop-gradient establishes the clean view as the ``reference'' and the degraded view as the ``student,'' preventing the model from trivially satisfying the constraint by making both views equally uncertain. This design is inspired by knowledge distillation~\cite{hinton2015distilling}, where the teacher (clean view) provides soft targets for the student (degraded view), but here both views share the same network.

We set $\lambda_f = 0.5$ and $\lambda_p = 0.1$ throughout all experiments. The degradation probability $p_{\text{deg}} = 0.5$ balances degraded and clean training samples, preventing overspecialization.

\textbf{Key properties.} (1) \textit{Zero additional parameters}---both views share the same backbone and classifier. (2) \textit{Zero inference overhead}---only one forward pass at test time, identical to baseline. (3) \textit{Complementary to augmentation}---DCPT does not replace augmentation but adds an orthogonal training signal. (4) \textit{Degradation-type agnostic at inference}---the model does not need to know the degradation type at test time.

\begin{figure*}[t]
\centerline{\includegraphics[width=0.95\textwidth]{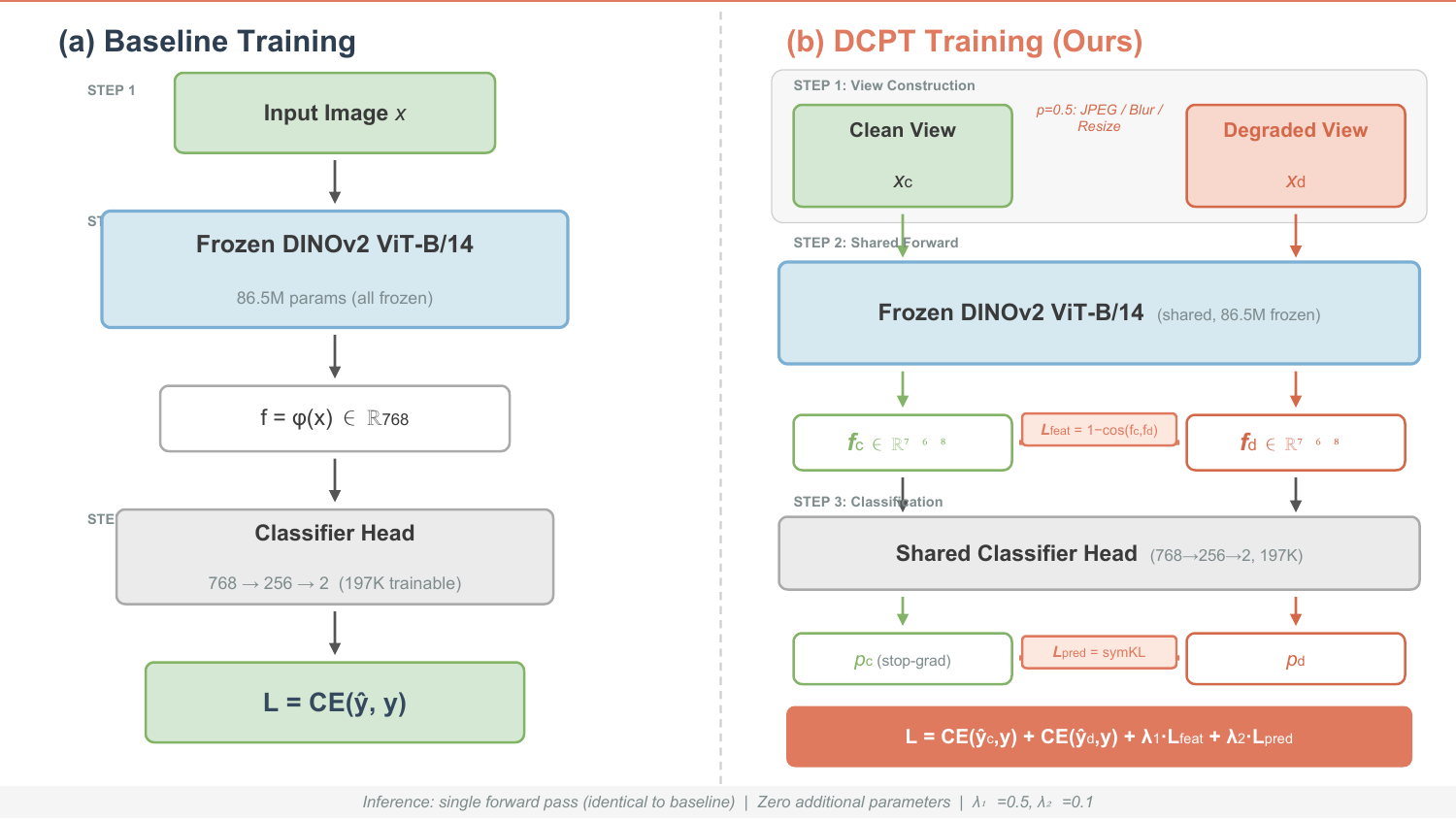}}
\caption{Comparison of (a) our baseline training and (b) DCPT training under the frozen-backbone setting. Both use the same frozen DINOv2 ViT-B/14 backbone and trainable classifier head (197K params). DCPT adds two consistency constraints ($\mathcal{L}_{\text{feat}}$ and $\mathcal{L}_{\text{pred}}$) between clean and degraded view representations, with zero additional parameters. Note: the original B-Free~\cite{guillaro2025bfree} fine-tunes the backbone end-to-end; our frozen setting is adopted to isolate the training objective contribution.}
\label{fig:arch}
\end{figure*}

\textbf{Overview.} Fig.~\ref{fig:arch} illustrates the comparison between baseline training and DCPT. The only difference is the loss function---the architecture remains identical.

% ══════════════════════════════════════════════════════════════
\section{Experiments}

\subsection{Experimental Setup}

\textbf{Training data.} We use the official B-Free training set~\cite{guillaro2025bfree}: 51,517 MS-COCO real images paired with 51,518 SD~2.1-generated fakes via self-conditioned inpainting.

\textbf{Evaluation.} Synthbuster contains 9,000 images from 9 generators (DALL-E 2/3, Firefly, GLIDE, Midjourney~v5, SD~1.3/1.4/2/XL), each with 1,000 images. We use 1,000 COCO real images as the real reference set.

\textbf{Degradation protocol.} JPEG compression (QF $\in \{30, 50, 70\}$), Gaussian blur ($\sigma \in \{1, 2, 3\}$), and bicubic downsampling (scale $\in \{0.5, 0.25\}$).

\textbf{Metrics.} Accuracy (ACC) as the primary metric, AUC (Area Under the ROC Curve) as a secondary metric.

\textbf{Implementation.} DINOv2 ViT-B/14 with registers is frozen. AdamW optimizer, lr=$10^{-4}$, weight decay=0.01, 20 epochs, batch size 64, single NVIDIA RTX 4090D (24GB), mixed precision training (AMP).

\textbf{Baselines.} (1) B-Free official~\cite{guillaro2025bfree}: the original model with end-to-end fine-tuned backbone (86.5M trainable params) and 5-crop inference---included as a reference upper bound under a fundamentally different (and much stronger) training regime. (2) Our baseline: frozen DINOv2 backbone + 197K classifier head, no paired training---the direct comparison target for DCPT under identical architecture.

\subsection{Degradation Robustness}

Table~\ref{tab:main} reports the average ACC (\%) across 9 Synthbuster generators. We include the B-Free official model~\cite{guillaro2025bfree} (end-to-end fine-tuned, 86.5M params, 5-crop inference) as a reference upper bound; note that it operates under a fundamentally different and much stronger training regime than our frozen-backbone experiments. \textit{Our primary comparison is between the frozen-backbone baseline and DCPT}, which share the identical architecture (197K trainable params, single-crop inference). The baseline achieves 86.2\% clean accuracy but degrades catastrophically under JPEG compression (56.8\% at QF=70, near random at QF=30). DCPT maintains 72.5\% under JPEG QF=70 and 69.5\% under QF=30---improvements of \textbf{+15.7\%} and \textbf{+17.2\%} respectively. The degraded average improves from 64.0\% to 73.1\% (\textbf{+9.1\%}), while clean accuracy drops by only 0.9\%.

\begin{table}[t]
\caption{Degradation robustness on Synthbuster (avg across 9 generators). Best within same-architecture in \textbf{bold}.}
\label{tab:main}
\begin{center}
\footnotesize
\setlength{\tabcolsep}{2pt}
\begin{tabular}{@{}llrcccccccccc@{}}
\toprule
\textbf{Method} & & \textbf{Prm} & \textbf{Cln} & \textbf{J70} & \textbf{J50} & \textbf{J30} & \textbf{B1} & \textbf{B2} & \textbf{B3} & \textbf{R.5} & \textbf{R.25} & \textbf{Avg} \\
\midrule
\multirow{2}{*}{B-Free$^\ast$\!\!\cite{guillaro2025bfree}} & ACC & 86.5M & 90.1 & 84.8 & 82.8 & 80.1 & 85.4 & 84.7 & 83.0 & 82.4 & 77.0 & 82.5 \\
 & AUC & & 98.7 & 97.4 & 96.3 & 94.8 & 98.5 & 97.7 & 95.5 & 98.0 & 92.9 & 96.4 \\
\midrule
\multirow{2}{*}{Baseline} & ACC & 197K & 86.2 & 56.8 & 53.3 & 52.3 & 85.3 & 77.2 & 52.3 & 77.0 & 57.7 & 64.0 \\
 & AUC & & 89.1 & 79.8 & 77.7 & 75.3 & 88.0 & 84.7 & 70.1 & 84.2 & 68.1 & 78.5 \\
\midrule
\multirow{2}{*}{\textbf{DCPT}} & ACC & 197K & \textbf{85.3} & \textbf{72.5} & \textbf{71.2} & \textbf{69.5} & \textbf{84.4} & \textbf{78.8} & \textbf{59.9} & \textbf{80.8} & \textbf{68.0} & \textbf{73.1} \\
 & AUC & & \textbf{87.9} & \textbf{80.5} & \textbf{79.4} & \textbf{77.6} & \textbf{87.1} & \textbf{84.0} & \textbf{71.2} & \textbf{83.8} & \textbf{71.4} & \textbf{79.4} \\
\bottomrule
\end{tabular}
\end{center}
\vspace{-4pt}
{\scriptsize $^\ast$Fully fine-tuned backbone + 5-crop. Avg = mean over 8 degraded conditions. J=JPEG QF, B=Blur $\sigma$, R=Resize scale. Baseline and DCPT use frozen backbone + single crop.\par}
\end{table}

\begin{figure}[t]
\centerline{\includegraphics[width=\columnwidth]{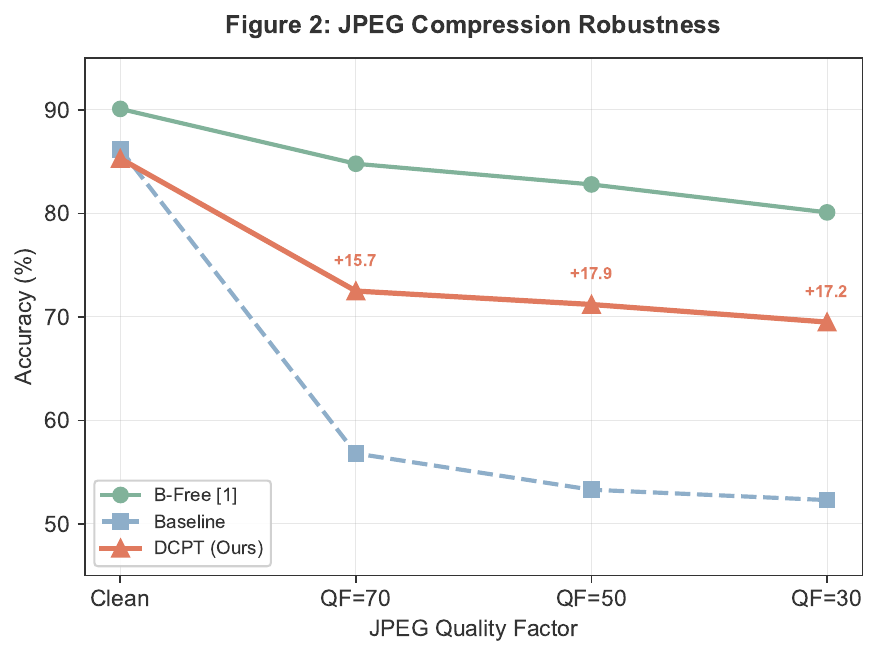}}
\caption{JPEG compression robustness. DCPT maintains a significantly flatter accuracy curve than the baseline, with improvements widening as compression increases.}
\label{fig:jpeg_curve}
\end{figure}

Fig.~\ref{fig:jpeg_curve} shows the accuracy as a function of JPEG quality factor (see also the architecture overview in Fig.~\ref{fig:arch}). The baseline drops precipitously from 86.2\% (clean) to 52.3\% (QF=30), approaching random chance. DCPT exhibits a much flatter profile: 85.3\% $\rightarrow$ 72.5\% $\rightarrow$ 69.5\%. The absolute improvement widens with degradation severity (+15.7\% at QF=70, +17.2\% at QF=30), confirming that explicit consistency training provides increasing benefit under harder conditions.

\begin{figure}[t]
\centerline{\includegraphics[width=\columnwidth]{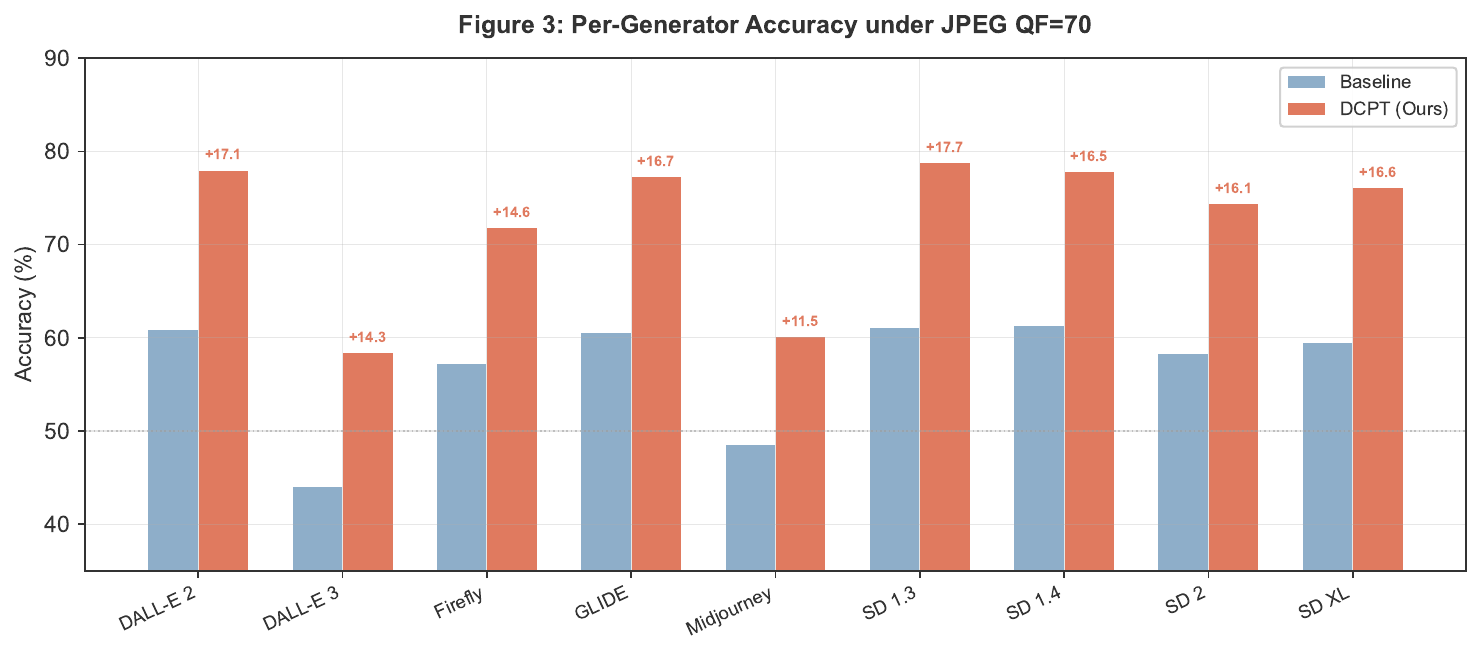}}
\caption{Per-generator accuracy under JPEG QF=70. DCPT consistently improves detection across all 9 generators.}
\label{fig:per_gen}
\end{figure}

Fig.~\ref{fig:per_gen} presents per-generator accuracy under JPEG QF=70. All 9 generators benefit from DCPT, with improvements ranging from +11.5\% (Midjourney~v5) to +17.7\% (SD~1.3). The improvement is consistent across both diffusion-based and GAN-based generators.

\subsection{Ablation Study}
\label{sec:ablation}

Table~\ref{tab:ablation} presents the ablation under fair comparison: all variants use the same frozen DINOv2 ViT-B/14 backbone.

\begin{table}[t]
\caption{Ablation of DCPT loss components (Synthbuster, JPEG conditions). All variants use frozen DINOv2 + classifier head (197K params unless noted).}
\label{tab:ablation}
\begin{center}
\small
\begin{tabular}{lccccccc}
\toprule
\textbf{Variant} & $\lambda_f$ & $\lambda_p$ & \textbf{Clean} & \textbf{J70} & \textbf{J50} & \textbf{J30} & \textbf{JPEG Avg} \\
\midrule
Baseline & 0 & 0 & \textbf{86.2} & 56.8 & 53.3 & 52.3 & 54.1 \\
$\mathcal{L}_{\text{feat}}$ only & 0.5 & 0 & 84.6 & 71.7 & 71.4 & 69.0 & 70.7 \\
$\mathcal{L}_{\text{pred}}$ only & 0 & 0.1 & 84.8 & \textbf{74.7} & \textbf{72.9} & \textbf{71.0} & \textbf{72.9} \\
\textbf{DCPT (both)} & 0.5 & 0.1 & 85.3 & 72.5 & 71.2 & 69.5 & 71.1 \\
\midrule
+ SRM (1.6M) & 0.5 & 0.1 & 78.1 & 50.7 & 56.4 & 65.8 & 57.6 \\
\bottomrule
\end{tabular}
\end{center}
\vspace{-4pt}
{\scriptsize J=JPEG QF. JPEG Avg = mean of J70/J50/J30. Full degradation results for DCPT in Table~\ref{tab:main} (Deg Avg=73.1, $\Delta$=+9.1).\par}
\end{table}

We decompose DCPT's two loss components under JPEG degradation (Table~\ref{tab:ablation}), the most challenging condition. Using $\mathcal{L}_{\text{feat}}$ alone ($\lambda_f{=}0.5, \lambda_p{=}0$) improves JPEG average from 54.1\% to 70.7\% (+16.6\%). $\mathcal{L}_{\text{pred}}$ alone ($\lambda_f{=}0, \lambda_p{=}0.1$) achieves the highest JPEG average of 72.9\% (+18.8\%), suggesting that prediction-level consistency via symmetric KL divergence provides a particularly strong training signal. Interestingly, their combination in full DCPT yields a slightly lower JPEG average (71.1\%) but the highest clean accuracy among DCPT variants (85.3\%), resulting in the best overall balance across all degradation types (Deg Avg=73.1, $\Delta$=+9.1; see Table~\ref{tab:main}). Adding an SRM frequency residual branch (+1.4M parameters) causes severe overfitting: training accuracy reaches 98.9\% while JPEG average drops to 57.6\%, worse than $\mathcal{L}_{\text{feat}}$ or $\mathcal{L}_{\text{pred}}$ alone. This confirms that with limited training data ($\sim$100K pairs), the training objective modification is more effective than architectural augmentation. Notably, while $\mathcal{L}_{\text{pred}}$ alone excels under JPEG, the full DCPT combination achieves the best overall degraded average (73.1\%) by also improving blur and resize conditions (Table~\ref{tab:main}: +7.6\% at blur $\sigma$=3, +10.3\% at resize 0.25), where feature-level consistency $\mathcal{L}_{\text{feat}}$ provides complementary regularization that stabilizes representations under spatial distortions.

\subsection{Analysis}

\textbf{Why JPEG is hardest.} JPEG compression introduces 8$\times$8 block artifacts that directly interfere with high-frequency generation fingerprints. Blur and resize, while reducing resolution, better preserve the global image structure captured by DINOv2 features. DCPT's consistency training helps most under JPEG because it forces the model to learn features that survive quantization.

\begin{figure}[t]
\centerline{\includegraphics[width=0.85\columnwidth]{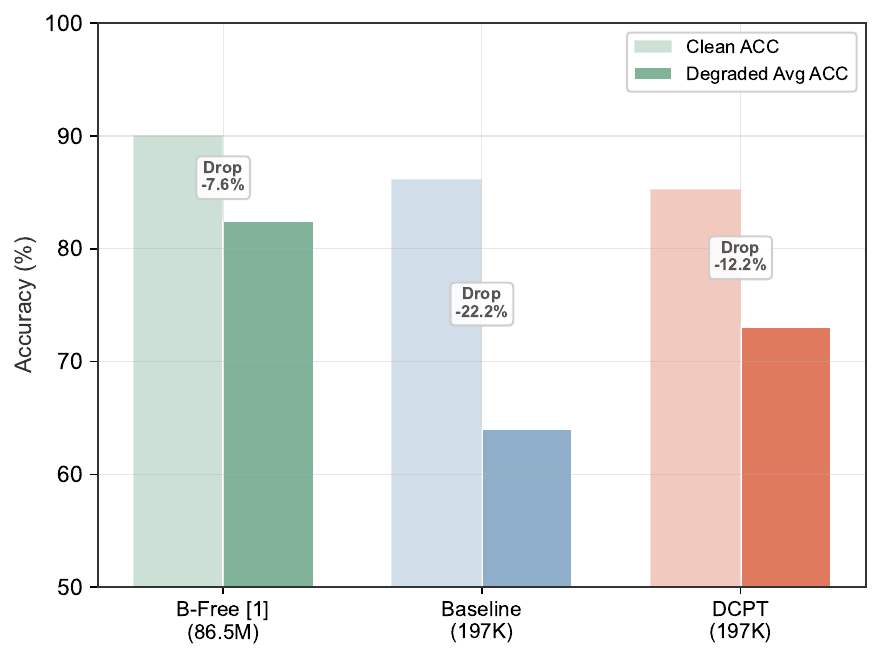}}
\caption{Clean vs.\ degraded accuracy comparison. DCPT nearly halves the clean-to-degraded performance gap.}
\label{fig:drop}
\end{figure}

\textbf{Clean-degraded trade-off.} Fig.~\ref{fig:drop} compares the clean-to-degraded accuracy drop across methods. The baseline suffers a 22.2 percentage point drop (86.2\% $\rightarrow$ 64.0\%), while DCPT reduces this to 12.2 points (85.3\% $\rightarrow$ 73.1\%)---nearly halving the degradation gap.

\begin{figure}[t]
\centerline{\includegraphics[width=\columnwidth]{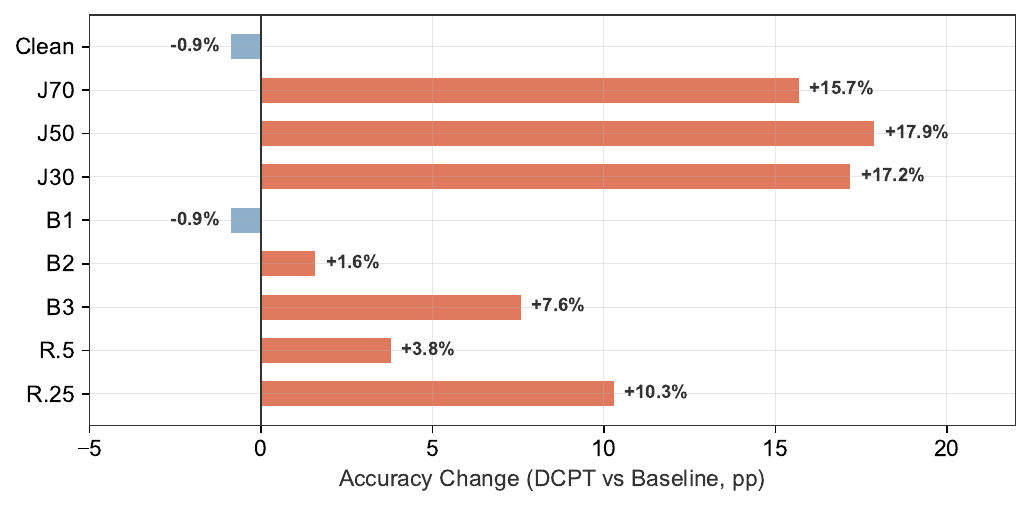}}
\caption{Per-degradation accuracy improvement of DCPT over baseline. JPEG compression benefits most.}
\label{fig:delta}
\end{figure}

\textbf{Per-degradation improvements.} Fig.~\ref{fig:delta} shows the improvement of DCPT over the baseline for each degradation type. JPEG benefits most (+15.7\% to +17.9\%), followed by resize (+3.8\% to +10.3\%) and blur (+1.6\% to +7.6\%). Mild blur ($\sigma$=1) and mild resize (0.5) show minimal change, as DINOv2 features are already somewhat robust to these mild corruptions.

% ══════════════════════════════════════════════════════════════
\section{Conclusion}

We presented Degradation-Consistent Paired Training (DCPT), a simple and effective training strategy for improving the degradation robustness of AI-generated image detectors. By constructing clean-degraded view pairs and enforcing feature and prediction consistency, DCPT explicitly optimizes for robustness without adding any parameters or inference cost. Experiments demonstrate a +9.1\% improvement in degraded-condition accuracy with only 0.9\% clean accuracy trade-off. Our ablation reveals that architectural additions cause overfitting on limited data, while the training objective modification delivers the core benefit. This finding has broader implications: in the era of frozen foundation models, \textit{how} we train matters more than \textit{what} we add.

Future work includes applying DCPT to fully fine-tuned models like B-Free, extending consistency constraints to unseen degradation types, and evaluating on the GenImage benchmark~\cite{zhu2023genimage} for cross-generator generalization under degradation.

\section*{Acknowledgment}

This work is supported by the Open Research Project of the State Key Laboratory of Industrial Control Technology, China (Grant No. ICT2025B70). Supported by Jiangxi Provincial Natural Science Foundation (Grant No. 20242BAB20041, Grant No. 20232BAB212006) and Hubei Provincial Natural Science Foundation of China (Grant No. 2023AFB474, Grant No. 2024AFB881). Supported by Anhui Provincial Special Project for Special Needs in Humanities and Social Sciences (Grant No. 2025AHGXSK50067) and Postgraduate Quality Engineering Project of Anhui Province (Grant No. 2024jyjxggyjY232).

% ══════════════════════════════════════════════════════════════

\end{document}